\pdfoutput=1

\documentclass[11pt]{article}
\usepackage{graphicx}
\usepackage{ACL2023}

\usepackage{forest}
\usetikzlibrary{shadows}
\definecolorseries{colours}{hsb}{grad}[hsb]{.575,1,1}{.987,-.234,0}
\resetcolorseries[12]{colours}
\usepackage{times}
\usepackage{amsmath}
\usepackage{latexsym}
 \usepackage{multirow}
 \usepackage{graphicx}
\usepackage{xcolor}
\usepackage{algorithm}
\usepackage{algorithmicx}
\usepackage{algpseudocode}

\usepackage[T1]{fontenc}
\usepackage{lipsum}

\usepackage[utf8]{inputenc}
\usepackage{microtype}

\usepackage{inconsolata}

\usepackage{lipsum} 
 \usepackage{graphicx}
 \usepackage[normalem]{ulem}
 \useunder{\uline}{\ul}{}
\usepackage{xcolor}
\definecolor{darkgreen}{rgb}{0.0, 0.5, 0.0}

\title{FinGrAct: A Framework for FINe-GRrained Evaluation of ACTionability in Explainable Automatic Fact-Checking}

\author{
  \textbf{Islam Eldifrawi},
  ~\textbf{Shengrui Wang},
  ~\textbf{Amine Trabelsi} \\
Department of Computer Science,
Université de Sherbrooke \\
    \normalsize{\tt{\{Islam.Eldifrawi;Shengrui.Wang;Amine.Trabelsi\}@usherbrooke.ca} }
}

\begin{document}
\graphicspath{ {./images/} }
\maketitle
\begin{abstract}

The field of explainable Automatic Fact-Checking (AFC) aims to enhance the transparency and trustworthiness of automated fact-verification systems by providing clear and comprehensible explanations. However, the effectiveness of these explanations depends on their actionability—their ability to empower users to make informed decisions and mitigate misinformation. Despite actionability being a critical property of high-quality explanations, no prior research has proposed a dedicated method to evaluate it. This paper introduces FinGrAct, a fine-grained evaluation framework that can access the web and it is designed to assess actionability in AFC explanations through well-defined criteria and an evaluation dataset. FinGrAct surpasses state-of-the-art (SOTA) evaluators, achieving the highest Pearson and Kendall correlation with human judgments while demonstrating the lowest ego-centric bias, making it a more robust evaluation approach for actionability evaluation in AFC.

\end{abstract}

\section{Introduction}


Explanation of claim veracity is essential in automated fact-checking as it enhances transparency, fosters trust, and educates users by clarifying why a claim is deemed true or not. It provides contextual understanding, and prevents the spread of misinformation. 

In the domain of explainable automated fact-checking (AFC), the explanation of a claim’s predicted veracity is assessed based on the presence of specific desired properties, referred to as "desiderata," as outlined by \citet{kotonya-toni-2020-explainable}. While some of these desiderata have been explored in the field of summarization, others, such as actionability, a critical property in fact-checking, remain unexplored. To date, no automatic evaluator for actionability has been developed. However, some general purpose SOTA evaluators emerged, but only for summarization tasks. In this work, these evaluators are adapted to AFC.


\begin{figure*}
    \centering
    \includegraphics[width=\textwidth]{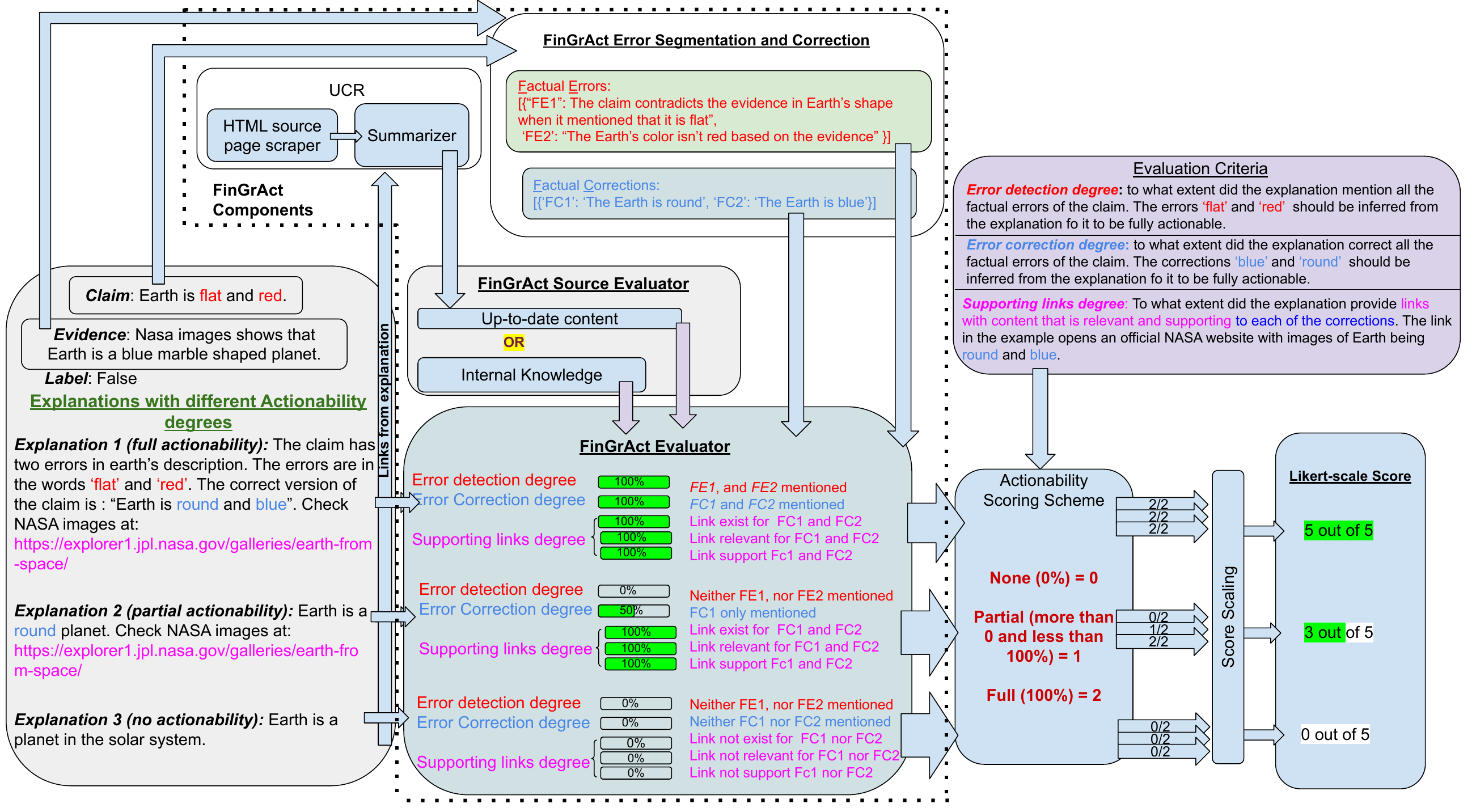}
    \caption{\textbf{FinGrAct Flow}. \textit{Input}: Claim, Evidence, Label, Explanation to be evaluated (with varying degrees of actionability or none). \textit{Processing}: FinGrAct Error Segmentation and Correction, FinGrAct Evaluator, FinGrAct Source Evaluator, URL Content Retriever (UCR). \textit{Output}: 3 evaluation scores (Error Detection, Error Correction, Supporting Links Degrees) classified as None (0), Partial (1), Full (2), aggregated into a Likert scale score (1 to 5).
   }
    \label{act_deg}
\end{figure*}

Actionability remains underexplored due to its complexity. According to \citet{kotonya-toni-2020-explainable}, actionability in AFC refers to \textbf{providing factual corrections for identified errors in a non-factual claim, supported by reliable sources and references}. This suggests that actionability is highly correlated with other key properties, such as relevance to the claim and completeness, where the explanation must provide a comprehensive context for why the claim is considered true or not. Evaluating actionability presents significant challenges as it requires well-defined criteria for judgment, as well as access to external sources in the internet to verify the reliability of supporting references. Figure \ref{act_deg} shows examples of explanations with different degrees of actionability.
To tackle the challenge of automatically evaluating the actionability of explanations in AFC systems, we introduce FinGrAct. FinGrAct is a fine-grained auto-evaluator that systematically measures the degree of actionability in these explanations. The evaluation framework is illustrated in Figure \ref{act_deg}.

Our contributions are as follows: 
\begin{enumerate}
    \item We present a dataset, constructed from existing sources, containing explanations for claim labels with varying degrees of actionability. Additionally, it includes human-rated actionability scores, which can serve as benchmarks for evaluating the performance of different evaluators of actionability.
    \item We present FinGrAct a novel fine-grained evaluator of explanations for actionability within the context of explainable AFC, based on LLM prompting, that correlates better with human ratings than other adapted SOTA evaluators. In addition, we showed that adding a simple component, named the URL Content Retriever (UCR), to fetch and evaluate web-link \textbf{textual} content, enhanced the performance of actionability evaluation in all SOTA evaluators. UCR enables LLMs to assess the validity and relevance of the supporting sources referenced in explanations.
    \item Furthermore, we conduct an investigation into the ego-centric bias present in LLMs. ego-centric bias happens when LLMs acting as judges to the output of various LLMs including themselves, tend to assign higher scores to their own output. In addition, FinGrAct has the least ego-centric bias among all other adapted SOTA evaluators.
\end{enumerate}
 
\section{Related Work}
\label{r_sur}

With the emergence of LLMs such as ChatGPT, recent studies have employed these models as evaluators \cite{fu2024gptscore} to assess LLM performance across various benchmarks, including 
G-EVAL \cite{liu2023g}, and FineSurE \cite{song2024finesure}. Nonetheless, it has been observed that LLMs functioning as auto-raters tend to show a preference for their own generated responses \cite{panickssery2024llm}. This is called ego-centric bias.

Training open-source general-purpose LLM autoraters has been investigated recently. TIGERScore, a Llama-2 model trained on GPT-4 generated error analysis data across multiple tasks, such as summarization, translation, and instruction-following, is presented by \citealt{jiang2023tigerscore}. However, it doesn't follow a likert-scale scoring or can it be scaled to it. For instance, it can produce a score of -12 making it hard to be compared to other evaluators. Prometheus \cite{kim2023prometheus}, InstructScore \cite{xu2023instructscore}, and Prometheus-2 \cite{kim2024prometheus} are comparable methods. \citealt{vu2024foundational} developed reward models used for aligning LLMs to human preferences via reinforcement learning and called their model FLAME. However, they haven't published their model or their dataset collection.

SOTA evaluators like G-Eval and Prometheus are widely used in summarization to assess properties such as coherence and faithfulness. Unlike property-specific tools like FineSurE, they are adaptable for evaluating new attributes. FineSurE gives a specific fine-grained definition for each property  it measures in a summary based on the key-facts it found in the transcript. Making its adaption to new tasks and new properties very hard.  

All the mentioned work is directed towards evaluating certain desired properties in diverse tasks like summarization and question answering. However, no effort has been directed to date towards developing an autorater for the desiderata of the explanations of AFC except when \citet{10.1016/j.engappai.2024.109492} tried to evaluate properties like (self)-contradiction, hallucination, convincingness and overall quality. However, they didn`t address the properties mentioned by \citet{kotonya-toni-2020-explainable, eldifrawi2024automated} that were deemed critical for fact-checking explanation like actionability for instance.
\section{FinGrAct Framework}
\subsection{Implementation Details}
\label{impl}
Inspired by recent advancements in fine-grained evaluation criteria for summarization—particularly in assessing key properties such as faithfulness and completeness, as demonstrated in works like \citet{ye2023flask, zhang2024qapyramid, song2024finesure}—we propose a specialized fine-grained evaluation methodology designed to assess the actionability of AFC explanations. Our approach, similar to these studies, leverages targeted prompting of specific LLMs to ensure a rigorous and nuanced assessment. 
%
As depicted in Figure \ref{act_deg}, we propose a divide-and-conquer approach to evaluating the actionability of an explanation—given a false claim and supporting evidence—by breaking it down into three distinct tasks: \textit{Error Segmentation and Correction}, \textit{Explanation Evaluation}, and \textit{Source Evaluation}.
We have the detailed prompts for every task in the Appendix \ref{FinGrAct_propmts}.

\textbf{Error Segmentation and Correction:}
This task is necessary for FinGrAct to preprocess a claim, extract what it identifies as \textit{factual errors}, and determine how \textit{corrections} within this claim should be made based on reliable given evidence (see Figure \ref{act_deg} for an example). Once this information is extracted, FinGrAct can evaluate the actionability of any given explanation of why a claim is false in the next task by aligning it with this information. Here, the underlying LLM is instructed to decompose the claim into atomic (sub)claims, assessing each for factual errors based on the evidence. It then explains the error (error reason) (e.g., FE1 in Figure \ref{act_deg}) and provides a subclaim correction (e.g., FC1 in Figure \ref{act_deg}). The output consists of triples (JSON output): false subclaim, error reason, and correction for each false subclaim. The corresponding prompt to this description is in Appendix \ref{FinGrAct_propmts}.

\textbf{Explanation Evaluation: FinGrAct Evaluator} 
Given lists of error explanations and generated corrections from the previous task, this step verifies whether these elements are explicitly inferred from the provided explanation. For a given explanation, the evaluation outputs a boolean value ``Yes" or ``No" for each error and each correction across all false subclaims.

The prompt for this evaluation phase also includes instructions to assess the web sources mentioned in the explanation. Their verification is conducted in the Sources Evaluation step.

\textbf{Sources Evaluation:} 
The goal here is to determine whether a link in an explanation exists, is relevant, and its content supports the needed corrections. Two methods were tested.
The first relies on the LLM’s prior knowledge, while the second involves retrieving the link’s content using an external component (Figure \ref{act_deg}), the \textit{URL Content Retriever (UCR)}.
For the first method (without UCR), the LLM is instructed to check, based on its knowledge, if there are relevant links in the explanation that support the needed corrections and to respond with a Yes or No.
For the second method (with UCR), 
the UCR external component is integrated \textit{before prompting} the evaluator to verify and validate the sources referenced in the explanations. The `requests' library in python is used to scrap the text of the web-page of the link in several steps:
Firstly, the HTML textual content of each link is scraped.
Secondly, all HTML tags are removed to extract clean, readable text.
Thirdly, the extracted textual content is then summarized using the MiniLM-L6-v2 model \cite{susanto2024performance}. Summarization is used to control the amount of tokens that will be inputted into the LLM and to discard any irrelevant text.
Lastly, the summarized content is subsequently incorporated into the prompt provided to the auto-evaluator, which assesses the relevance of the link's content and determines whether it supports the needed corrections. More specifically, the LLM is instructed to check from the output of the UCR if the links are working (Yes or No for Link exist, Figure \ref{act_deg}), to check if the content is relevant (Yes or No, Link relevant) and to check if it supports the corrections (Yes or No, Link support).


For a given explanation on why a claim is false, the output of this phase is a set of false subclaims $S$, each containing information on whether the explanation has mentioned the error, corrected the error, and includes a functional, relevant, or supporting link or not (for the version without UCR, only error mention, error correction and supporting link exist).

\subsection{Actionability Scoring Scheme}
\label{act_scr}
\begin{algorithm}[H]
\caption{Actionability Scoring Algorithm (Case With UCR)}
\label{alg:actionability_scoring}
\begin{algorithmic}[1]
\Require False subclaims $S = \{s_1, s_2, \dots, s_n\}$
\Ensure Scores $(E_d, E_c, L_e, L_r, L_s)$

\State $E_d \gets 0, E_c \gets 0, L_e \gets 0, L_r \gets 0, L_s \gets 0$
\For{$s_i \in S$}
    \State $E_d \gets E_d + \mathbf{1}(s_i$ has detected errors$)$
    \State $E_c \gets E_c + \mathbf{1}(s_i$ has corrected errors$)$
    \State $L_e \gets L_e + \mathbf{1}(s_i$ has a functional link$)$
    \State $L_r \gets L_r + \mathbf{1}(s_i$ has a relevant link$)$
    \State $L_s \gets L_s + \mathbf{1}(s_i$ has a supporting link$)$
\EndFor
\State \Comment{Categorize (i): 2 if i = 1, 1 if 0 < i < 1, else 0}
\State $E_d \gets \text{Categorize}(E_d/n)$
\State $E_c \gets \text{Categorize}(E_c/n)$
\State $L_e \gets \frac{\text{Categorize}(L_e/n)}{2}$
\State $L_r \gets \frac{\text{Categorize}(L_r/n)}{4}$
\State $L_s \gets \frac{\text{Categorize}(L_s/n)}{4}$
\State \Return $(E_d, E_c, L_e + L_r + L_s)$
\end{algorithmic}
\end{algorithm}
%
Given the output described in the previous paragraph, this phase deals with returning a final evaluation score of the actionability of an explanation.
As shown in Algorithm \ref{alg:actionability_scoring} and Figure \ref{act_deg}, error detection and correction are categorized into three levels: full, partial, or none. \textbf{\textit{Error Detection:}} A score of 2 is awarded if all factual errors in the claim are fully identified, 1 if only some errors are detected, and 0 if no errors are recognized (Categorize function in Algorithm \ref{alg:actionability_scoring}). \textbf{\textit{Error Correction:}} If all identified errors are fully corrected, the explanation receives a score of 2; if only some are addressed, it scores 1; and if no corrections are made, it scores 0. \textbf{\textit{Supporting Links:}} Evaluation is based on the sum of the scores of three key factors: Link functionality – whether the link is accessible (score: 1). Relevance – whether the linked content pertains to the explanation's topic (score: 0.5). Support – whether the linked content directly substantiates the corrections needed to make the claim true. (score: 0.5).

The maximum possible score is 6. To normalize this score for comparison with other state-of-the-art (SOTA) evaluators, we apply a scaling factor of 5/6, approximating the final score to a Likert scale ranging from 0 to 5.


\section{The Evaluation Dataset}
\label{data}
We collected 203 examples from two existing benchmark datasets, ensuring a diverse range of error detection and correction levels.
\subsection{Data Collection and creation}
The objective is to develop a dataset that encompasses varying levels of actionability. 
Consequently, the dataset must include explanations with different degrees of error detection and correction, with some incorporating supporting references while others do not. Furthermore, a distinct category of actionable explanations includes counterfactual (CF) explanations. In conclusion, the dataset should include both counterfactual and non-counterfactual explanations, each demonstrating different levels of actionability.

The sources of this dataset are two benchmark datasets. The first was created by \citealt{10.1145/3534678.3539205}, and it contains false claims and counterfactual explanations that explains the reason why the claims are false in three different formats. From this dataset, we were able to generate six different categories of actionable explanations as shown in Figure \ref{fig:img0}. The categories are:  
\textbf{Error Detection Only:} The explanation only highlights the error in the claim.
\textbf{Error Correction Only:} The explanation only provides a corrected version of the non-factual claim.
\textbf{Error Correction and Detection Only:} The explanation does both error detection and correction, however, it doesn't provide any sources/links that support its content.
\textbf{Error Detection with Sources:} The explanation does error detection and it has sources/links that should support its content.
\textbf{Error Correction with Sources:} The explanation does error correction and it has sources/links that should support its content.
\textbf{Error Detection and Correction with Sources:} The explanation does error detection and correction, and it has sources that should support its content.

The second source is the data from \cite{kotonya-toni-2020-explainable-automated}. The explanations are a form of summarization of the evidence and they also contain different degrees of actionability. The addition of this dataset guarantees that the actionable explanations will not be only presented by Counter-factual explanations. We extracted two categories from this dataset, false and partially true claims with their evidence and explanations. Half of each category has sources supporting it and half doesn't, resulting in four different categories in Figure \ref{fig:img0}. 

\subsection{Dataset Creation}
\begin{figure*}
    \centering
    \includegraphics[width=\textwidth]{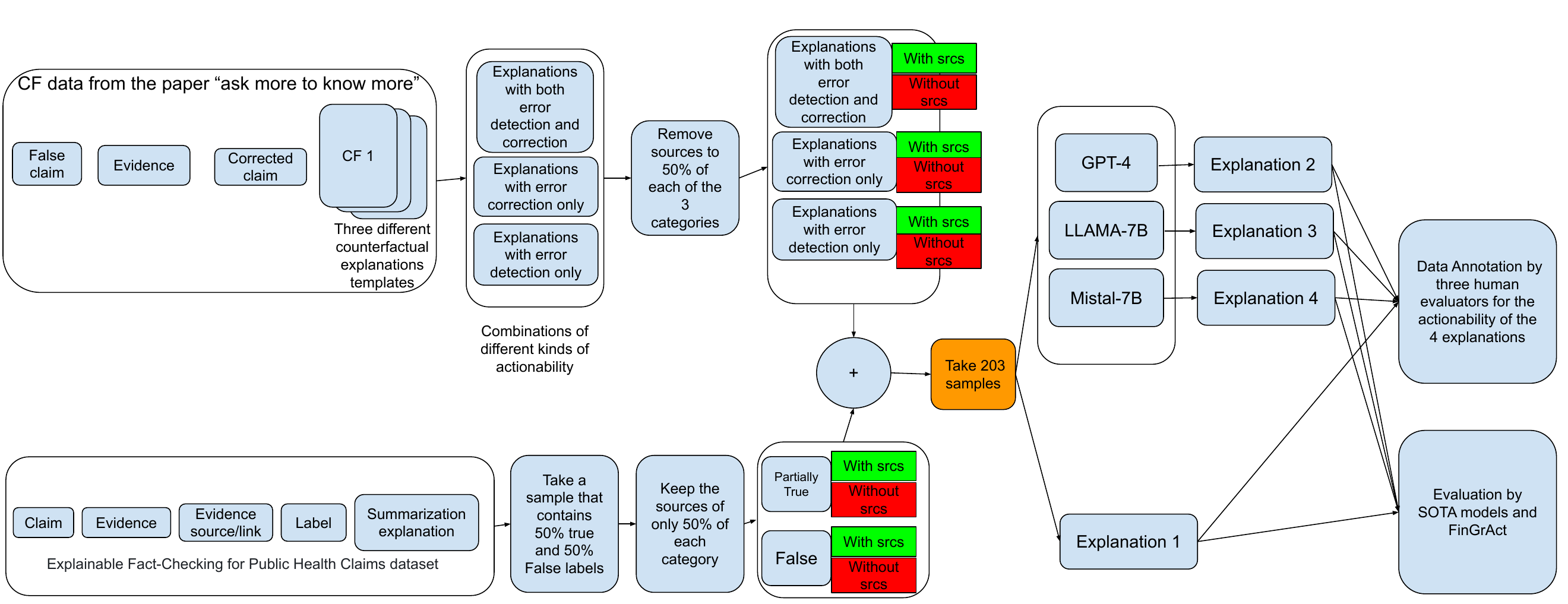}
    \caption{\textbf{The dataset creation process}.The dataset combines counterfactual data from \citealt{10.1145/3534678.3539205}, categorized by error detection, correction, and supporting sources, with non-counterfactual actionable explanations from \cite{kotonya-toni-2020-explainable-automated}. The combined dataset was then sampled and used to generate three additional explanations from three LLMs to analyze ego-centric bias.}
    \label{fig:img0}
\end{figure*}

The dataset was constructed in three stages. The first, illustrated in Figure \ref{fig:img0}, involves categorizing the counterfactual (CF) data from \cite{10.1145/3534678.3539205} into six distinct categories. These categories, detailed in Section \ref{data}, are based on the extent of error detection, the level of correction provided, and the presence of supporting sources.

Then the second as shown in Figure \ref{fig:img0} is where the dataset from \cite{kotonya-toni-2020-explainable-automated} is used to generate explanations that are divided into four different categories as mentioned in Section \ref{act_scr}.  

The data generated from both stages are merged and then sampled to obtain 203 random instances, ensuring representation across all categories mentioned earlier in Section \ref{act_scr}. As a result, we have constructed a diverse dataset that includes varying degrees of actionability, comprising both counterfactual (CF) and non-counterfactual explanations. 
Subsequently, the dataset is augmented with other generated explanations from three large language models (LLMs): LLAMA-7B, Mistral-7B and GPT-4. These generative models serve as the foundation for Prometheus, G-Eval, and FinGrAct. The generated explanations are only later utilized in the ego-centric bias study to assess whether the evaluators exhibit preferential bias toward outputs generated by their respective underlying LLMs.

\begin{figure}
    \centering
    \includegraphics[width=6cm]{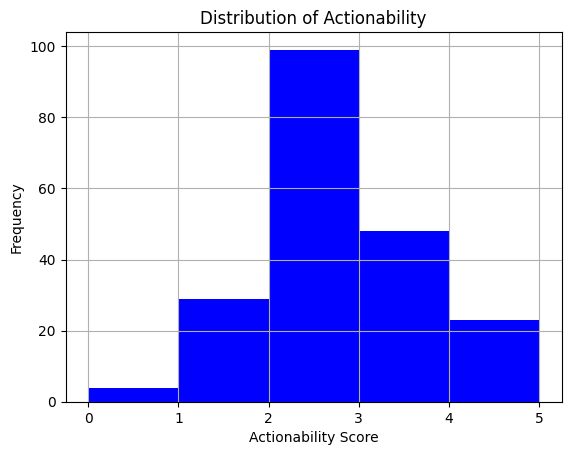}
    \caption{Distribution of actionability in the dataset based on average human annotation rating. This distribution shows that the dataset has diverse degrees of actionability.}
    \label{fig:img4}
\end{figure}

Finally, each claim is accompanied by four explanations—one sourced from the combined dataset and three generated by the specified LLMs, following the prompts detailed in the Appendix \ref{gen}. 
Three human annotators independently assess the actionability of each explanation based on the provided evaluation guidelines, also outlined in Appendix \ref{ann}. Additionally, the annotators underwent training through a video demonstration and followed an iterative annotation process. Initially, they were provided with a small subset of the data and encouraged to ask questions and share feedback. Based on their input, the instructions were refined to minimize confusion and clarify the task. This process was repeated in stages, with annotators receiving additional data incrementally, ensuring continuous improvement in understanding and consistency in annotation. The scores assigned by the annotators are then averaged and normalized to fit a Likert scale from 0 to 5, ensuring consistency with the scoring system used by the evaluators. The dataset shows diverse degrees of actionability based on the human annotation scoring distribution (Figure \ref{fig:img4}).

\section{Experimentation and Results}
We applied our fine-grained evaluation methodology for actionability using OpenAI’s GPT-4-1106-preview model. This choice was made because GPT-4-1106-preview serves as the primary model for G-Eval and other baseline evaluators, ensuring a fair and consistent comparison between FinGrAct and existing evaluators such as G-Eval. The focus of our study is not on the model itself but rather the evaluation methodology, ensuring that the assessment framework remains the central point of analysis. We design three distinct experiments:

\textbf{Comparison with SOTA Evaluation Methods:} The first experiment aims to compare our evaluation methodology with existing state-of-the-art (SOTA) models. The primary metric used for comparison is the correlation with human annotations.

\textbf{Testing the External URL Content Retriever Component:} The second experiment assesses the effectiveness of our external component, which aids in retrieving and processing link content. We analyze its impact on the correlation with human annotations across different models and evaluation methodologies.

\textbf{Ego-Centric Bias Analysis}: The third experiment investigates ego-centric bias and how our methodology influences this bias, which was initially identified in G-Eval by \citet{liu2023g}.

In all experiments, the reported scores represent the rounded average of three independent runs for each model. This approach ensures a degree of consistency in the results, reducing variability and improving the reliability of our findings. In the next subsections, the baseline models, setup and results of each experiment will be discussed.

\subsection{Baseline Models}
In the summarization domain, zero-shot state-of-the-art (SOTA) LLM-based evaluators such as G-Eval and Prometheus are widely used to assess key properties such as coherence, faithfulness, and completeness, among others. 
G-Eval and Prometheus are general-purpose and can be adapted to evaluate new properties and they can be used in new tasks. This adaptability makes them particularly valuable as SOTA baseline evaluators. Since these baseline models were never used to evaluate actionability, the adaptation prompts are in Appendix \ref{geval_adap} for G-Eval, and in Appendix \ref{prom_adap} for Prometheus. 

\subsection{Experiment 1: Comparison with SOTA Evaluation Methods}
\label{exp1}
\textbf{Setup:} To evaluate FinGrAct against Prometheus and G-Eval, human annotators' scores serve as the ground truth benchmark. The performance of each evaluation methodology is measured using two correlation metrics: Pearson correlation and Kendall’s tau correlation with human scores. The original parameter settings for both G-Eval and Prometheus are preserved to ensure a fair comparison:
G-Eval: temperature = 1, n = 20, top\_p = 1
Prometheus: temperature = 1, top\_p = 0.9
 For FinGrAct, we set the temperature to zero and clear the conversation history before processing each new sample in GPT-4, following best practices from prior research \cite{shen-etal-2023-large, song-etal-2024-finesure} to enhance reproducibility.

In this setup, the links and sources within the explanations are evaluated without access to the web. Instead, LLMs assess the credibility and relevance of these sources solely based on their pre-existing, and potentially outdated, knowledge. This means that the evaluation of sources relies on the model’s internal representations rather than real-time verification.

\textbf{Results:} In this experiment, as shown in Table \ref{tab:table1}, the Pearson correlation with human evaluations indicates that G-Eval achieves 0.147, Prometheus scores 0.328, and FinGrAct attains 0.46. Notably, FinGrAct correlates better than Prometheus (13.2\% greater Pearson Correlation) demonstrating its superiority over the adapted SOTA evaluators in aligning with human judgments. 

Similarly, for Kendall’s tau correlation, G-Eval achieves 0.117, Prometheus scores 0.294, and FinGrAct reaches 0.409. Again, FinGrAct outperforms Prometheus by 11.5\%, reinforcing its effectiveness in producing evaluations that better correlate with human assessments. 

\textbf{Analysis:} It is noteworthy that LLMs tend to achieve higher correlation with human annotations when provided with more detailed instructions. This could explain why Prometheus exhibits a higher correlation with human evaluations compared to G-Eval, as Prometheus requires a detailed scoring rubric and a clear definition of the property being evaluated, whereas G-Eval relies solely on the property definition. 

FinGrAct, in contrast, takes a more structured approach by dividing the evaluation into three tasks, each further broken down into subtasks, thereby providing even more detailed guidance on the actionability evaluation process than both Prometheus and G-Eval. This granular methodology likely contributes to FinGrAct achieving the highest correlation with human annotations.

\begin{table}[]
\centering
\resizebox{\columnwidth}{!}{%
\begin{tabular}{c|clclcl|clclcl|}
\cline{2-13}
                              & \multicolumn{6}{c|}{Without UCR}                                                                                                                                                     & \multicolumn{6}{c|}{With UCR}                                                                                                                                                        \\ \cline{2-13} 
                              & \multicolumn{2}{c|}{Geval} & \multicolumn{2}{c|}{\begin{tabular}[c]{@{}c@{}}Prome-\\ theus\end{tabular}} & \multicolumn{2}{c|}{\begin{tabular}[c]{@{}c@{}}FinGr-\\ Act\end{tabular}} & \multicolumn{2}{c|}{Geval} & \multicolumn{2}{c|}{\begin{tabular}[c]{@{}c@{}}Prome-\\ theus\end{tabular}} & \multicolumn{2}{c|}{\begin{tabular}[c]{@{}c@{}}FinGr-\\ Act\end{tabular}} \\ \hline
\multicolumn{1}{|c|}{Pearson} & \multicolumn{2}{c|}{0.147} & \multicolumn{2}{c|}{0.328}                                                  & \multicolumn{2}{c|}{{\ul 0.460}}                                          & \multicolumn{2}{c|}{0.213} & \multicolumn{2}{c|}{0.405}                                                  & \multicolumn{2}{c|}{\textbf{0.520}}                                       \\ \hline
\multicolumn{1}{|c|}{Kendall} & \multicolumn{2}{c|}{0.117} & \multicolumn{2}{c|}{0.294}                                                  & \multicolumn{2}{c|}{{\ul 0.409}}                                          & \multicolumn{2}{c|}{0.207} & \multicolumn{2}{c|}{0.341}                                                  & \multicolumn{2}{c|}{\textbf{0.419}}                                       \\ \hline
\end{tabular}%
}
\caption{The following table presents the Pearson and Kendall Tau correlations between human annotator and SOTA evaluators for explanations in the combined dataset. The first half of the table displays the correlation values without incorporating the URL textual content retriever (UCR), while the second half shows the correlations after its inclusion. The underlined scores are the highest scores without incorporating the UCR, while the bold scores are those after incorporating it. All the p-values are less than 0.05.}
\label{tab:table1}
\end{table}

\subsection{Experiment 2: Testing the External URL Content Retrieval Component}
\textbf{Setup:} The same setup outlined in Section \ref{exp1} is used, with one key modification: before constructing the prompt, the links are processed. The source pages of the links are scraped, their text is extracted, and then summarized using the UCR Section \ref{impl}.

Next, the summarized content from all linked sources is concatenated and incorporated into the prompt. The LLM is then tasked with evaluating whether the aggregated content from these URLs is both relevant to and supportive of the explanation and the needed corrections to the claim. Finally, the actionability score is computed as mentioned in Section \ref{act_scr}. 

\textbf{Results:} As shown in Table \ref{tab:table1}, incorporating the UCR —which retrieves and integrates the actual content of the linked sources into the prompt instead of relying solely on the LLM’s internal knowledge—led to an increase in correlation with human annotations across all SOTA evaluators in both Pearson and Kendall’s tau correlations. G-Eval’s Pearson correlation increased from 0.147 to 0.213, reflecting a 6.6\% improvement, while its Kendall’s tau correlation rose from 0.117 to 0.207, a 9\% increase. Similarly, Prometheus showed a 7.7\% improvement in Pearson correlation and a 4.7\% increase in Kendall’s tau correlation. Lastly, FinGrAct’s Pearson correlation improved by 6\%, while its Kendall’s tau correlation saw a 1\% increase. 

\textbf{Analysis:} These results demonstrate that providing real, up-to-date source content enhances evaluation accuracy for assessing actionability. However, more improvement was expected after adding the UCR as it should provide real-time web content.  Upon further investigation, the primary reason for this limited improvement is that some linked pages contain only images or JavaScript-rendered content, which the UCR cannot process as it is not multi-modal. When no textual content is extracted, the LLM assumes that the URL’s content is neither relevant nor supportive, leading to a lower actionability score. Meanwhile, human annotators can interpret images and JavaScript-based elements, recognize their relevance, and assign higher scores, creating a discrepancy between LLM-based and human evaluations.

\subsection{Experiment 3: Ego-Centric Bias Analysis}

\begin{table}[]
\resizebox{\columnwidth}{!}{%
\begin{tabular}{c|c|c|c|}
\cline{2-4}
                                                                                                                                 & Geval & Promethuse & FinGrAct \\ \hline
\multicolumn{1}{|c|}{\begin{tabular}[c]{@{}c@{}}\# of scores > human\\ scores + 2\end{tabular}} & 99    & 52         & 17     \\ \hline
\multicolumn{1}{|c|}{\begin{tabular}[c]{@{}c@{}}\# of scores < human\\ scores - 2\end{tabular}}   & 12    & 10         & 4      \\ \hline
\end{tabular}%
}
\caption{The analysis examines ego-centric bias in evaluator scoring across 203 samples. It identifies cases where an evaluator overestimates actionability by scoring at least 2 units higher than human annotations and instances where it underestimates actionability, with human scores exceeding the evaluator's by 2 or more units. This assessment highlights discrepancies between model evaluations and human judgment.}
\label{tab2}
\end{table}

In their study, \citet{liu2023g} identified a bias in evaluators, where they tend to favor their own model's generations over those from other models, even when the latter are objectively better. \citet{ohi-etal-2024-likelihood} introduced a method for detecting this bias, which they termed 'Likelihood-based Evaluation Bias.' However, this approach requires access to the probability distribution of the LLM's generations, which is often unavailable, especially when working with commercial LLMs. \citet{ye2024justice} also addressed this issue, referring to it as 'ego-centric Bias,' and we adopt this terminology in our work. Their research primarily focuses on understanding the effects of this bias on performance and strategies for mitigating it. The purpose of this study is to compare the effect of ego-centric bias on our fine-grained evaluation "FinGrAct" and on other SOTA evaluators. In this paper, we propose a simple yet effective method for identifying this bias within the context of actionability evaluation in explainable AFC when the probability distribution of LLM generations is not available.

We identify biased samples by observing that evaluators tend to assign significantly higher scores to explanations generated by their own underlying LLMs compared to human ratings. For instance, Geval exhibits a preference for GPT-4-generated explanations, even when alternative explanations may be superior. To quantify this bias, we implement a Likert-scale scoring system ranging from 0 to 5, allowing for a tolerance of a 1-point difference between human and LLM scores. If multiple human annotators rate an explanation as 2 and the LLM assigns a 3, the sample is excluded from bias analysis. However, if the LLM scores the same explanation as 4 or 5, it is classified as ego-centric bias. Thus, a discrepancy of 2 or more points higher than the human rating serves as a clear indicator of bias.

\textbf{Setup:} To measure ego-centric bias, each evaluator is tasked with assessing explanations generated by its underlying LLM. For instance, Prometheus evaluates Mistral-generated explanations, while G-Eval and FinGrAct evaluate GPT-4-generated explanations. Their evaluations are then compared against human annotations, and instances of bias are identified and counted. Specifically, cases where an evaluator overestimates actionability—assigning a score at least 2 units higher than human ratings—are classified as ego-centric bias. 

The scores from the three human annotators were averaged and compared against the mean scores from three evaluation runs for each automatic evaluator. Geval exhibited the highest variance, with 113 biased samples in the first run, 101 in the second, and 84 in the third, averaging 99. Prometheus demonstrated more stability, with 55 biased samples in the first run, 50 in the second, and 53 in the third. FinGrAct showed the least variance, with 17 biased samples in the first run, 19 in the second, and 17 in the third.

Additionally, instances where the evaluators underestimate actionability relative to human judgments are also recorded. This analysis allows us to determine whether ego-centric bias or underestimation contributes more to the misalignment between automated evaluators and human assessments

\textbf{Results:} Out of 203 samples, the results clearly indicate that ego-centric bias contributes significantly more to the misalignment between human annotations and LLM-based evaluations than underestimation does. G-Eval exhibits ego-centric bias in 99 out of 203 samples (48.7\%), whereas underestimation occurs in only 12 samples (5.9\%). Similarly, Prometheus demonstrates bias in 26\% of cases, while underestimation accounts for just 5\%. FinGrAct, which shows the least bias, has 8.4\% biased samples and 2\% underestimation. 

\textbf{Analysis:} These findings suggest that LLM-based evaluators tend to overestimate actionability far more frequently than they underestimate it, highlighting a key limitation in their judgment alignment with human evaluations. It is worth noting that LLMs as evaluators tend to exhibit higher correlation with human annotations and lower ego-centric bias when the evaluation criteria are more detailed. In G-Eval, the evaluation relies primarily on the definition of the property being measured, accompanied by general instructions, but without structured guidance as shown in Figure \ref{geval_prompt} in Appendix \ref{geval_adap}. Prometheus improves upon this by incorporating a detailed scoring rubric, which provides more explicit evaluation guidelines (as shown in the Prometheus prompt Figure \ref{promethuse_prompt} in Appendix \ref{prom_adap}). FinGrAct, however, employs the most structured approach, implementing a detailed framework where claims are segmented, errors are explicitly identified, and corrections are validated along with supporting links, each scored in granular detail as mentioned in Appendix \ref{FinGrAct_propmts}. This meticulous evaluation process explains why FinGrAct exhibits the lowest ego-centric bias and the fewest mis-alignments due to underestimation.

\section{Conclusion}

The paper introduces FinGrAct, a fine-grained evaluation method for assessing actionability in AFC, a crucial but underexplored property. Actionability is essential for explainable AFC systems but remains challenging to evaluate. The study shows that FinGrAct outperforms SOTA evaluators, achieving the highest Pearson and Kendall correlation with human judgments and exhibiting the lowest ego-centric bias, making it more reliable. Additionally, incorporating retrieved and summarized content from referenced sources further improved actionability evaluation across all models. These findings establish FinGrAct as a superior framework for assessing actionability in AFC.

\section*{Limitations}
The limitations can be summarized in the following points:
\begin{enumerate}
    \item The URL content retriever (UCR) component is currently limited to extracting textual content from the provided URLs. This restriction led to performance issues in instances where the referenced URLs contained primarily images or JavaScript-based content, as no retrievable text was available. As a result, these cases were misinterpreted, affecting the overall evaluation accuracy.In future work, we plan to develop a multimodal URL content retriever capable of processing both textual and non-textual content, including images and JavaScript-rendered elements. This enhancement will ensure more comprehensive content retrieval, leading to a more accurate and reliable evaluation.
    \item All experiments in this study were conducted using a zero-shot learning approach. Fine-tuning was not explored due to its high computational and financial cost, particularly when applied to commercial LLMs like GPT-4. Future work may consider cost-effective fine-tuning strategies or alternative methods to enhance evaluation performance without incurring significant resource demands.
    \item Our study primarily focused on ego-centric bias in LLM-based evaluations. However, in future work, we plan to explore other types of biases, including cross-model biases, where different LLMs may exhibit preferential treatment toward explanations generated by certain other LLMs. This broader analysis will provide a more comprehensive understanding of biases in LLM-based evaluation systems and help develop fairer and more reliable evaluation methodologies.  
\end{enumerate}

 \section*{Ethics Statement}
 This paper focuses on designing an automatic evaluator utilizing LLMs for explaniable AFC using a combination two benchmark datasets. Some errors might be induced in the evaluation of explanations as the evaluator is LLM based.

\bibliography{anthology,custom}

\begin{thebibliography}{21}
\expandafter\ifx\csname natexlab\endcsname\relax\def\natexlab#1{#1}\fi

\bibitem[{Dai et~al.(2022)Dai, Hsu, Xiong, and Ku}]{10.1145/3534678.3539205}
Shih-Chieh Dai, Yi-Li Hsu, Aiping Xiong, and Lun-Wei Ku. 2022.
\newblock \href {https://doi.org/10.1145/3534678.3539205} {Ask to know more: Generating counterfactual explanations for fake claims}.
\newblock In \emph{Proceedings of the 28th ACM SIGKDD Conference on Knowledge Discovery and Data Mining}, KDD '22, page 2800–2810, New York, NY, USA. Association for Computing Machinery.

\bibitem[{Eldifrawi et~al.(2024)Eldifrawi, Wang, and Trabelsi}]{eldifrawi2024automated}
Islam Eldifrawi, Shengrui Wang, and Amine Trabelsi. 2024.
\newblock Automated justification production for claim veracity in fact checking: A survey on architectures and approaches.
\newblock In \emph{Proceedings of the 62nd Annual Meeting of the Association for Computational Linguistics (Volume 1: Long Papers)}, pages 6679--6692.

\bibitem[{Feher et~al.(2025)Feher, Khered, Zhang, Batista-Navarro, and Schlegel}]{10.1016/j.engappai.2024.109492}
Darius Feher, Abdullah Khered, Hao Zhang, Riza Batista-Navarro, and Viktor Schlegel. 2025.
\newblock \href {https://doi.org/10.1016/j.engappai.2024.109492} {Learning to generate and evaluate fact-checking explanations with transformers}.
\newblock \emph{Eng. Appl. Artif. Intell.}, 139(PA).

\bibitem[{Fu et~al.(2024)Fu, Ng, Jiang, and Liu}]{fu2024gptscore}
Jinlan Fu, See~Kiong Ng, Zhengbao Jiang, and Pengfei Liu. 2024.
\newblock Gptscore: Evaluate as you desire.
\newblock In \emph{Proceedings of the 2024 Conference of the North American Chapter of the Association for Computational Linguistics: Human Language Technologies (Volume 1: Long Papers)}, pages 6556--6576.

\bibitem[{Jiang et~al.(2023)Jiang, Li, Zhang, Huang, Lin, and Chen}]{jiang2023tigerscore}
Dongfu Jiang, Yishan Li, Ge~Zhang, Wenhao Huang, Bill~Yuchen Lin, and Wenhu Chen. 2023.
\newblock Tigerscore: Towards building explainable metric for all text generation tasks.
\newblock \emph{Transactions on Machine Learning Research}.

\bibitem[{Kim et~al.(2023)Kim, Shin, Cho, Jang, Longpre, Lee, Yun, Shin, Kim, Thorne et~al.}]{kim2023prometheus}
Seungone Kim, Jamin Shin, Yejin Cho, Joel Jang, Shayne Longpre, Hwaran Lee, Sangdoo Yun, Seongjin Shin, Sungdong Kim, James Thorne, et~al. 2023.
\newblock Prometheus: Inducing fine-grained evaluation capability in language models, 2024.
\newblock \emph{URL https://arxiv. org/abs/2310.08491}.

\bibitem[{Kim et~al.(2024)Kim, Suk, Longpre, Lin, Shin, Welleck, Neubig, Lee, Lee, and Seo}]{kim2024prometheus}
Seungone Kim, Juyoung Suk, Shayne Longpre, Bill~Yuchen Lin, Jamin Shin, Sean Welleck, Graham Neubig, Moontae Lee, Kyungjae Lee, and Minjoon Seo. 2024.
\newblock Prometheus 2: An open source language model specialized in evaluating other language models.
\newblock \emph{arXiv preprint arXiv:2405.01535}.

\bibitem[{Kotonya and Toni(2020{\natexlab{a}})}]{kotonya-toni-2020-explainable}
Neema Kotonya and Francesca Toni. 2020{\natexlab{a}}.
\newblock \href {https://doi.org/10.18653/v1/2020.coling-main.474} {Explainable automated fact-checking: A survey}.
\newblock In \emph{Proceedings of the 28th International Conference on Computational Linguistics}, pages 5430--5443, Barcelona, Spain (Online). International Committee on Computational Linguistics.

\bibitem[{Kotonya and Toni(2020{\natexlab{b}})}]{kotonya-toni-2020-explainable-automated}
Neema Kotonya and Francesca Toni. 2020{\natexlab{b}}.
\newblock \href {https://doi.org/10.18653/v1/2020.emnlp-main.623} {Explainable automated fact-checking for public health claims}.
\newblock In \emph{Proceedings of the 2020 Conference on Empirical Methods in Natural Language Processing (EMNLP)}, pages 7740--7754, Online. Association for Computational Linguistics.

\bibitem[{Liu et~al.(2023)Liu, Iter, Xu, Wang, Xu, and Zhu}]{liu2023g}
Yang Liu, Dan Iter, Yichong Xu, Shuohang Wang, Ruochen Xu, and Chenguang Zhu. 2023.
\newblock G-eval: Nlg evaluation using gpt-4 with better human alignment.
\newblock In \emph{Proceedings of the 2023 Conference on Empirical Methods in Natural Language Processing}, pages 2511--2522.

\bibitem[{Ohi et~al.(2024)Ohi, Kaneko, Koike, Loem, and Okazaki}]{ohi-etal-2024-likelihood}
Masanari Ohi, Masahiro Kaneko, Ryuto Koike, Mengsay Loem, and Naoaki Okazaki. 2024.
\newblock \href {https://doi.org/10.18653/v1/2024.findings-acl.193} {Likelihood-based mitigation of evaluation bias in large language models}.
\newblock In \emph{Findings of the Association for Computational Linguistics: ACL 2024}, pages 3237--3245, Bangkok, Thailand. Association for Computational Linguistics.

\bibitem[{Panickssery et~al.(2024)Panickssery, Bowman, and Feng}]{panickssery2024llm}
Arjun Panickssery, Samuel~R Bowman, and Shi Feng. 2024.
\newblock Llm evaluators recognize and favor their own generations.
\newblock \emph{arXiv e-prints}, pages arXiv--2404.

\bibitem[{Shen et~al.(2023)Shen, Cheng, Nguyen, You, and Bing}]{shen-etal-2023-large}
Chenhui Shen, Liying Cheng, Xuan-Phi Nguyen, Yang You, and Lidong Bing. 2023.
\newblock \href {https://doi.org/10.18653/v1/2023.findings-emnlp.278} {Large language models are not yet human-level evaluators for abstractive summarization}.
\newblock In \emph{Findings of the Association for Computational Linguistics: EMNLP 2023}, pages 4215--4233, Singapore. Association for Computational Linguistics.

\bibitem[{Song et~al.(2024{\natexlab{a}})Song, Su, Shalyminov, Cai, and Mansour}]{song2024finesure}
Hwanjun Song, Hang Su, Igor Shalyminov, Jason Cai, and Saab Mansour. 2024{\natexlab{a}}.
\newblock Finesure: Fine-grained summarization evaluation using llms.
\newblock In \emph{Proceedings of the 62nd Annual Meeting of the Association for Computational Linguistics (Volume 1: Long Papers)}, pages 906--922.

\bibitem[{Song et~al.(2024{\natexlab{b}})Song, Su, Shalyminov, Cai, and Mansour}]{song-etal-2024-finesure}
Hwanjun Song, Hang Su, Igor Shalyminov, Jason Cai, and Saab Mansour. 2024{\natexlab{b}}.
\newblock \href {https://doi.org/10.18653/v1/2024.acl-long.51} {{F}ine{S}ur{E}: Fine-grained summarization evaluation using {LLM}s}.
\newblock In \emph{Proceedings of the 62nd Annual Meeting of the Association for Computational Linguistics (Volume 1: Long Papers)}, pages 906--922, Bangkok, Thailand. Association for Computational Linguistics.

\bibitem[{Susanto et~al.(2024)Susanto, Ferdiana, and Adji}]{susanto2024performance}
Budi Susanto, Ridi Ferdiana, and Teguh~Bharata Adji. 2024.
\newblock Performance of traditional and dense vector information retrieval models in code search.
\newblock In \emph{2024 2nd International Conference on Software Engineering and Information Technology (ICoSEIT)}, pages 52--57. IEEE.

\bibitem[{Vu et~al.(2024)Vu, Krishna, Alzubi, Tar, Faruqui, and Sung}]{vu2024foundational}
Tu~Vu, Kalpesh Krishna, Salaheddin Alzubi, Chris Tar, Manaal Faruqui, and Yun-Hsuan Sung. 2024.
\newblock Foundational autoraters: Taming large language models for better automatic evaluation.
\newblock In \emph{Proceedings of the 2024 Conference on Empirical Methods in Natural Language Processing}, pages 17086--17105.

\bibitem[{Xu et~al.(2023)Xu, Wang, Pan, Song, Freitag, Wang, and Li}]{xu2023instructscore}
Wenda Xu, Danqing Wang, Liangming Pan, Zhenqiao Song, Markus Freitag, William Wang, and Lei Li. 2023.
\newblock Instructscore: Towards explainable text generation evaluation with automatic feedback.
\newblock In \emph{Proceedings of the 2023 Conference on Empirical Methods in Natural Language Processing}, pages 5967--5994.

\bibitem[{Ye et~al.(2024)Ye, Wang, Huang, Chen, Zhang, Moniz, Gao, Geyer, Huang, Chen et~al.}]{ye2024justice}
Jiayi Ye, Yanbo Wang, Yue Huang, Dongping Chen, Qihui Zhang, Nuno Moniz, Tian Gao, Werner Geyer, Chao Huang, Pin-Yu Chen, et~al. 2024.
\newblock Justice or prejudice? quantifying biases in llm-as-a-judge.
\newblock In \emph{Neurips Safe Generative AI Workshop 2024}.

\bibitem[{Ye et~al.(2023)Ye, Kim, Kim, Hwang, Kim, Jo, Thorne, Kim, and Seo}]{ye2023flask}
Seonghyeon Ye, Doyoung Kim, Sungdong Kim, Hyeonbin Hwang, Seungone Kim, Yongrae Jo, James Thorne, Juho Kim, and Minjoon Seo. 2023.
\newblock Flask: Fine-grained language model evaluation based on alignment skill sets.
\newblock \emph{arXiv preprint arXiv:2307.10928}.

\bibitem[{Zhang et~al.(2024)Zhang, Wan, Cattan, Klein, Dagan, and Bansal}]{zhang2024qapyramid}
Shiyue Zhang, David Wan, Arie Cattan, Ayal Klein, Ido Dagan, and Mohit Bansal. 2024.
\newblock Qapyramid: Fine-grained evaluation of content selection for text summarization.
\newblock \emph{arXiv e-prints}, pages arXiv--2412.

\end{thebibliography}
\bibliographystyle{acl_natbib}

\appendix
\section{Human annotation details}
\label{ann}
Three human annotators, MSc students familiar with NLP tasks, aged between 22 and 30 years, were tasked with evaluating the actionability of 203 samples, each containing four explanations. They were provided with detailed instructions, clear examples and a video demo to ensure consistency in their evaluations (see Figure \ref{ann_instruct}). 

The annotation process followed an iterative approach: initially, a subset of the dataset was assigned for evaluation, and annotators provided feedback on any ambiguities. Based on their input, the instructions were refined and improved to enhance clarity. Significant disagreements were addressed through discussions and successive refinements of the guidelines, ensuring a more consistent and reliable evaluation process.


We used Krippendorff’s alpha to measure inter-annotator agreement, as it is well-suited for Likert-scale data with multiple annotators. The resulting Krippendorff’s alpha of 0.863 reflects the high level of agreement, which is expected due to the following factors:
\begin{enumerate}
    \item \textbf{Clear Annotation Guidelines} – annotators have strict, well-defined rules. In addition, the process is iterative, fine-grained, and a video demonstration is provided to the annotators.
    \item \textbf{Objective or Easy-to-Classify Data} – Tasks with minimal ambiguity (e.g., labeling with '0,1 or 2' labels like "none, partial and full") often lead to high agreement.
    \item \textbf{Trained Annotators} tend to agree more than crowd-sourced or untrained annotators.
\end{enumerate}

The annotators volunteered to evaluate the actionability of the explanations for the claims. Given this, we can assert that their annotations were conducted solely based on their understanding of the provided instructions.
\begin{figure*}
\fcolorbox{blue}{white}{%
        \begin{minipage}{\textwidth}
\scriptsize
You have a claim that needs to be fact-checked, evidence which is basically the trustworthy information that we can rely on to check if the claim is factual or not, a label that shows the prediction of our model on whether the claim is true or false, and the model explanation for its predicted label. If the claim aligns with the evidence, then its label will be true and vice versa.

The explanation should be evaluated based on the following criteria:
\begin{enumerate}
    \item The number of factual errors that were detected in the claim and pointed out in the explanation compared to the number of all the factual errors on a scale from 0 to 2. This means that if no factual errors in the claim were detected then the score is zero, if some of the factual errors were detected then the score is one, and if \textbf{all} the factual errors were detected then the score is 2.
    \item The number of factual errors that were corrected faithfully in the explanation on a scale from 0 to 2. This means that if no factual errors in the claim were corrected then the score is zero, if some of the factual errors were corrected then the score is one, and if \textbf{all} the factual errors were corrected then the score is 2.
    \item The correctness of the resources/references, their relevance to the evidence, and their alignment with the evidence on a scale from 0 to 3. This means if there were no references mentioned then the score is zero,  if the references are mentioned and the links are correct then the score is one,if the references are mentioned and are relevant to the evidence then the score is 2 and if the references are mentioned and are \textbf{both relevant and aligned with the evidence} then it gets a score of 3.
\end{enumerate}
Example 1:

Claim:
Adrienne Bailon is an accountant.

Evidence:
Adrienne Eliza Houghton Bailon ; born October 24 , 1983 is an American singer-songwriter , recording artist , actress , dancer and television personality .

Label:
FALSE

Explanation:
Adrienne Bailon is an American singer-songwriter, recording artist, actress, dancer and television personality.
Evaluation:
\begin{itemize}
    \item percentage of factual errors detected in explanation: 0 out of 2 “no factual errors from the claim are mentioned explicitly in the explanation”
    \item percentage of factual errors corrected in explanation: 2 out of 2 “The error ‘accountant’ in the claim was corrected in the explanation’
    \item The correctness of the resources/references and their relevance: 0 out of 3 No references were mentioned
\end{itemize}

Example 2:

Claim:
Junun is a book.

Evidence:
Junun is a 2015 documentary film directed by Paul Thomas Anderson . Junun premièred at the 2015 New York Film Festival and was released on the MUBI film streaming service on October 9 and on iTunes on November 20, 2015 . Greenwood previously composed soundtracks for several Anderson films.

Label:
FALSE

Explanation:
If we were to say 'Junun is a 2015 documentary film directed by paul thomas anderson' instead of 'book', the claim would be correct.
https:\/\/en.wikipedia.org\/wiki\/Junun\_(film)

\begin{itemize}
    \item percentage of factual errors detected in explanation: 2 out of 2“ factual errors from the claim are mentioned in the explanation after the part ‘instead of’ ” and the error is explicitly obvious in the word “book”
    \item percentage of factual errors corrected in explanation: 2 out of 2“The error ‘book’ in the claim was corrected in the explanation to the word “film”
    \item The correctness of the resources/references and their relevance: 3 out of 3 “references mentioned are correct, aligned and relevant as I have opened the link and checked”
\end{itemize}

Example 3:

Claim:
Earth is flat and green.

Evidence:
Nasa images show that Earth is a huge blue round planet.

Label:
FALSE

Explanation:
The NASA images show that the Earth is huge and not green.
https:\/\/science.nasa.gov\/gallery\/jupiter 

Evaluation:
\begin{itemize}
    \item percentage of factual errors detected in explanation: 1 out of 2  “ factual errors from the claim are mentioned in the explanation in part ‘not green’ ”. However, the error of Earth being ‘flat’ wasn’t mentioned
    \item percentage of factual errors corrected in explanation: 1 out of 2 “The error ‘green’ in the claim was corrected in the explanation, but it didn’t say that it was blue. The other error ‘flat’ was not corrected
    \item The correctness of the resources\/references and their relevance:1 out of 3  “references mentioned are correct but not relevant as I have opened the link and found out that it shows Jupiter, not Earth”
\end{itemize}

 \end{minipage}%
}
    \caption{The instructions given to the three human annotators for evaluating the actionability}
    \label{ann_instruct}

\end{figure*}

\section{Explanations generated by the LLMs of the SOTA evaluators}
\label{gen}
To study ego-centric bias, we prompted the generative LLMs that serve as the foundation for our state-of-the-art (SOTA) evaluators to generate actionable explanations. These explanations are then assessed by both the same SOTA evaluators and human annotators, enabling a comparative analysis to detect potential biases.

This process consists of two steps to ensure diversity in explanations, including both those with and without supporting links. 
\begin{enumerate}
    \item We generate explanations for all claims using the prompt shown in Figure \ref{gen_propmt}.
    \item We prompt the LLMs to generate supporting links for \textbf{some} claims while leaving other explanations without links. We use the prompt in Figure \ref{lnk_gen}. This approach ensures a balanced dataset with explanations both with and without links.
\end{enumerate}

\begin{figure}
\fcolorbox{blue}{lime}{%
        \begin{minipage}{\columnwidth}
\scriptsize
You will be given a claim, some credible information called the evidence, a label that shows whether the claim is true or false.\\\\

Your task is to generate an actionable explanation for the label of the claim based on the evidence.\\\\

Please make sure you read and understand these instructions carefully. Please keep this document open while reviewing, and refer to it as needed.\\\\

Explanation Criteria:\\

\textbf{Actionability - misinformation detection and factual correction backed up by credible sources. The explanation should provide an indication of which parts of the claim include misalignment with the evidence. In addition, the explanation should provide a corrected version of the erroneous claim.}\\\\

Evaluation Steps:\\\\

1. Read the claim, evidence and the explanation carefully.\\
2. Compare the claim with the evidence and identify the errors or misalignment parts between the claim and the evidence.\\
3. Generate an explanation that clearly mentions the errors detected in the claim and corrects these errors based on the evidence.\\
4. Don't respond with any information outside the provided evidence. Your are restricted to answer from the evidence only.\\\\

Claim:\\
\{claim\}
\\\\
Evidence:\\
\{evidence\}
\\\\
Label:\\
\{label\}
 \end{minipage}%
}
    \caption{General prompt used on all the generative LLMs of the SOTA evaluators to generate actionable explanations. These explanations are used to measure the ego-centric bias of evaluators based on these underlying LLMs.}
    \label{gen_propmt}

\end{figure}

\begin{figure}
\fcolorbox{blue}{lime}{%
        \begin{minipage}{\columnwidth}
\scriptsize
You will be given some information called the explanation.\\

Your task is to generate a correct and working web link for a source supporting the explanation.\\

Please make sure you read and understand these instructions carefully. Please keep this document open while reviewing, and refer to it as needed.\\\\

Steps:\\

1. The web link provided should be correct and working.\\
2. The web link should open a page that has information relevant to the explanation.\\
3. The web link should open a page that has information supporting and in alignment with the explanation.\\\\

Explanation:\\
\{explanation\}
 \end{minipage}%
}
    \caption{General prompt used on all the generative LLMs of the SOTA evaluators to generate supporting links to their explanations for some samples}
    \label{lnk_gen}

\end{figure}

\section{G-Eval Adaptation to Measure Actionability}
\label{geval_adap}
G-Eval is widely used to assess various important properties in the summarization domain, but it has never been applied to measure actionability. This is because actionability is critical in explainable fact-checking, rather than in summarization. 

Typically, G-Eval takes a transcript and a summary as inputs for evaluation. However, in this work, we adapt it for actionability assessment by changing the inputs to the claim, evidence, label, and the explanation to be evaluated.

The customizability of G-Eval makes it a go-to evaluator and baseline for researchers. Its concept is straightforward: provide the LLM with the definition of the property to be evaluated, give it general guidelines or instructions for the evaluation process, and then let the LLM act as a judge, determining the score based on the given definition and instructions.

This approach gives the LLM significant flexibility in its assessments. As a result, the strength of the LLM-as-a-judge plays a crucial role in the quality and reliability of G-Eval's evaluations.

We used two prompts with G-Eval:

\begin{enumerate}
    \item \textbf{Actionability Evaluation Prompt} – This prompt includes a clear definition of actionability along with detailed evaluation instructions. This prompt is shown in Figure \ref{geval_prompt}.
    \item \textbf{Actionability Evaluation with URL Content Retrieval} – This prompt is identical to the first but incorporates the retrieved web content from all links mentioned in the explanation. This ensures that the evaluation considers external sources rather than relying solely on the LLM's prior knowledge. This prompt is shown in Figure \ref{geval_prompt_URL}.
\end{enumerate}

\begin{figure}
\fcolorbox{blue}{lime}{%
        \begin{minipage}{\columnwidth}
\scriptsize
You will be given a claim, some credible information called the evidence, a label that shows whether the claim is true or false, and an explanation for the label.
\\\\
Your task is to evaluate the explanation on one metric.
\\\\
Please make sure you read and understand these instructions carefully. Please keep this document open while reviewing, and refer to it as needed.
\\\\
Evaluation Criteria:
\\\\
\textbf{Actionability (0-5) - misinformation detection and factual correction backed up by credible sources. The explanation should provide an indication of which parts of the claim include misalignment with the evidence.In addition, the explanation should provide a corrected version of the erroneous claim.}
\\\\
Evaluation Steps:
\\\\
1. Read the claim, evidence and the explanations carefully.

2. Compare the claim with the evidence and identify the errors or misalignment parts between the claim and the evidence.

3. Assess how well the explanations cover the errors detected, the supporting sources provided, and the degree of error correction of the claim in each explanation.

4.If there is a link in the explanation, assess if the content of the link -based on your knowledge- is both relevant and supporting to the explanation.

5. Assign a score from 1 to 5 to the metric.

6. Your output should be scores only.
\\\\

Claim:

\{claim\}
\\\\
Evidence:

\{evidence\}
\\\\
Label:

\{label\}
\\

Explanation:

\{explanation\}
\\

Evaluation Form (scores ONLY):\\
- Actionability:
 \end{minipage}%
}
    \caption{Geval Prompt for evaluating actionability}
    \label{geval_prompt}

\end{figure}

\begin{figure}
\fcolorbox{blue}{lime}{%
        \begin{minipage}{\columnwidth}
\scriptsize
You will be given a claim, some credible information called the evidence, a label that shows whether the claim is true or false, and an explanation for
the label.
The explanation might have a link. If this is true, the content of the link will be provided as well.\\\\
Your task is to evaluate the explanation on one metric.
\\\\
Please make sure you read and understand these instructions carefully. Please keep this document open while reviewing, and refer to it as needed.
\\\\
Evaluation Criteria:
\\\\
\textbf{Actionability (0-5) - misinformation detection and factual correction backed up by credible sources.
The explanation should provide an indication of which parts of the claim include misalignment with the evidence.In addition, the explanation should provide a corrected version of the erroneous claim.}
\\\\
Evaluation Steps:
\\\\
1. Read the claim, evidence and the explanation carefully.\\
2. Compare the claim with the evidence and identify the errors or misalignment parts between the claim and the evidence.\\
3. Assess how well the explanation covers the errors detected and the degree of error correction of the claim in the explanation.\\
4. If there is a link in the explanation, assess if the content of the link provided is both relevant and supporting to the explanation.\\
5. Assign a score from 1 to 5 to the metric.\\
6. Your output should be scores only.
\\\\
Claim:

\{claim\}
\\\\
Evidence:

\{evidence\}
\\\\
Label:

\{label\}
\\\\
Explanation:

\{explanation\}
\\\\
link content:

\{link\_content\}
\\\\
Evaluation Form (scores ONLY):

- Actionability:
 \end{minipage}%
}
    \caption{Geval Prompt for evaluating actionability with URL content retriever}
    \label{geval_prompt_URL}

\end{figure}

\section{Prometheus Adaptation to Measure Actionability}
\label{prom_adap}
Prometheus is designed to assess customized properties in a transcript, but it has never been applied to measure actionability. It utilizes Mistral-7B and follows a strictly structured prompt template, where the evaluator inputs the definition of the property to be assessed, along with explicit instructions for evaluating that property—similar to G-Eval.

What differentiates Prometheus is its rigid scoring framework. Unlike G-Eval, which allows the LLM more freedom in judgment, Prometheus requires a detailed scoring rubric, specifying exactly when to assign a score of 1, 2, 3, 4, or 5. Additionally, it requests feedback to provide insights into the reasoning behind the assigned score.

Typically, Prometheus evaluates summaries by taking a transcript, a summary, and a detailed scoring rubric as input. In this work, however, we adapt it for actionability assessment by modifying the inputs to include the claim, evidence, label, explanation to be evaluated, and a structured scoring rubric.

This approach reduces the LLM’s flexibility in scoring but ensures greater consistency and reliability. As a result, Mistral-7B as an LLM-as-a-judge has outperformed G-Eval (GPT-4) in several benchmarks, making Prometheus the current state-of-the-art (SOTA) evaluator.

We used two prompts with Prometheus:

\begin{enumerate}
    \item \textbf{Actionability Evaluation Prompt} – This prompt includes a clear definition of actionability along with detailed evaluation instructions and scoring ruberic. This prompt is shown in Figure \ref{promethuse_prompt}.
    \item \textbf{Actionability Evaluation with URL Content Retrieval} – This prompt is identical to the first but incorporates the retrieved web content from all links mentioned in the explanation. This ensures that the evaluation considers external sources rather than relying solely on the LLM's prior knowledge. This prompt is shown in Figure \ref{promethuse_prompt_with_lnk}. 
\end{enumerate}

\begin{figure}

\fcolorbox{blue}{lime}{%
        \begin{minipage}{\columnwidth}
\scriptsize
An instruction (might include an Input inside it), a response to evaluate, and a score rubric representing a evaluation criteria are given.\\
1. Write a detailed feedback that assess the quality of the response strictly based on the given score rubric, not evaluating in general.\\
2. After writing a feedback, write a score that is an integer between 1 and 5. You should refer to the score rubric.\\
3. The output format should look as follows: \"Feedback: (write a feedback for criteria) [RESULT] (an integer number between 1 and 5)\"\\
4. Please do not generate any other opening, closing, and explanations.\\\\

\#\#\#The instruction to evaluate:
You will be given a claim, some credible information called the evidence, a label that shows whether the claim is true or false, and an response that explains the label.\\
\textbf{Evaluate the actionability of the response by examining misinformation detection and factual correction backed up by credible sources. The response should provide an indication of which parts of the claim include misalignment with the evidence. In addition, the response should provide a corrected version of the erroneous claim and a web link or a source that it relies on.}
\\\\
\#\#\#Claim:\\
\{claim\}
\\\\
\#\#\#Evidence:\\
\{evidence\}
\\\\
\#\#\#Label:\\
\{label\}
\\\\
\#\#\#Response to evaluate:\\
\{response\}
\\\\
\#\#\#Score Rubrics:\\\\
"criteria":"Is the model proficient in detecting and correcting the error or misalgnment between the response and the evidence and also providing supporting sources",\\\\
  "\textbf{score1\_description}":"The model detects the error or misalihnment without correcting it. In addition sources are not mentioned",\\
  "\textbf{score2\_description}":"The model corrects the error or misalignment, but doesn't point out where the error is. In addition sources are not mentioned",\\
  "\textbf{score3\_description}":"The model typically detects the error or misalignment and explicitly mentions it. The model also  provides correction of the error. In addition sources are not mentioned ",\\
  "\textbf{score4\_description}":"The model consistently detects the error or misalignment and explicitly mentions it. The model also  provides correction of the error. In addition sources are  mentioned",\\
  "\textbf{score5\_description}":"The model excels in the detection of errors or misalignment and explicitly mentions it. The model also  provides correction of the error. In addition,  sources/links that have relevant and supporting content are included in the explanation."

 \end{minipage}%
}
    \caption{Prometheus prompt for evaluating actionability}
    \label{promethuse_prompt}

\end{figure}

\begin{figure}
\fcolorbox{blue}{lime}{%
        \begin{minipage}{\columnwidth}
\scriptsize
\#\#\#Task Description:\\\\
An instruction (might include an Input inside it), a response to evaluate, and a score rubric representing a evaluation criteria are given.\\
1. Write a detailed feedback that assess the quality of the response strictly based on the given score rubric, not evaluating in general.\\
2. After writing a feedback, write a score that is an integer between 1 and 5. You should refer to the score rubric.\\
3. The output format should look as follows: \"Feedback: (write a feedback for criteria) [RESULT] (an integer number between 1 and 5)\"\\
4. Please do not generate any other opening, closing, and explanations.\\\\

\#\#\#The instruction to evaluate:\\\\
You will be given a claim, some credible information called the evidence, a label that shows whether the claim is true or false, and an response that explains the label.
If the response contains a link, the content of the link will be provided as well.
Evaluate the actionability of the response by examining misinformation detection and factual correction backed up by credible sources. The response should provide an indication of which parts of the claim include misalignment with the evidence. In addition, the response should provide a corrected version of the erroneous claim and a web link or a source that it relies on. The content of the link should support the response.
\\\\
\#\#\#Claim:\\
\{claim\}
\\\\
\#\#\#Evidence:\\
\{evidence\}
\\\\
\#\#\#Label:\\
\{label\}
\\\\
\#\#\#Response to evaluate:\\
\{response\}
\\\\
\#\#\#link content:\\
\{link\_content\}
\\\\
\#\#\#Score Rubrics:\\\\
"criteria":"Is the model proficient in detecting and correcting the error or misalgnment between the response and the evidence and also providing supporting sources",\\
  \textbf{"score1\_description"}:"The model detects the error or misalihnment without correcting it. In addition sources are not mentioned",\\
  \textbf{"score2\_description"}:"The model corrects the error or misalignment, but doesn't point out where the error is. In addition sources are not mentioned",\\
  \textbf{"score3\_description"}:"The model typically detects the error or misalignment and explicitly mentions it. The model also  provides correction of the error. In addition sources are not mentioned ",\\
  \textbf{"score4\_description}":"The model consistently detects the error or misalignment and explicitly mentions it. The model also  provides correction of the error. In addition sources are  mentioned",\\
  \textbf{"score5\_description"}:"The model excels in the detection of errors or misalignment and explicitly mentions it. The model also  provides correction of the error. In addition credible and faithful sources are mentioned. The content of the sources or links supports the response."
 \end{minipage}%
}
    \caption{Prometheus prompt for evaluating actionability after adding the URL content retriever}
    \label{promethuse_prompt_with_lnk}

\end{figure}

\section{FinGrAct and Fine-Grained Actionability Evaluation}
\label{FinGrAct_propmts}
The systematic and fine-grained evaluation approach in FinGrAct minimizes GPT-4’s reliance on its own knowledge by breaking down the evaluation into small, manageable tasks. This structured methodology ensures a more objective and transparent assessment of actionability.

FinGrAct's Error Segmentation operates in three stages:

\begin{enumerate}
    \item \textbf{Claim Breakdown} – The claim is broken down into atomic claims, making the evaluation more granular and precise.
    \item \textbf{Error Detection} – Each atomic claim is examined for factual inaccuracies.
    \item \textbf{Correction Proposal} – For every detected error, a correction is generated, ensuring that the explanation provides accurate and actionable insights.
\end{enumerate}

By structuring the evaluation in this way, FinGrAct reduces subjectivity and enhances the reliability of actionability assessments.

Here is an example:

\textbf{claim:} Earth is flat and red.

\textbf{Evidence:} Nasa images shows that Eart is a blue marble shaped planet. 

\textbf{Explanation:} The claim has two errors in earth’s description. The errors are in the words ‘flat’ and ‘red’. The correct version of the claim is : “Earth is round and blue”. Check NASA images at: Check NASA images at https://explorer1.jpl.nasa.gov/galleries/earth-from-space

\textbf{output of the error segmentation and correction} stage:
[
\{'sentence': 'Earth is flat', 'reason': 'The evidence explicitly states that Earth is a marble shaped planet, not flat.', 'correction': 'Earth is round.'\}, \{'sentence': 'Earth is red', 'reason': 'As per the evidence, Earth is blue, 'correction': 'Earth is blue'\}
]

The prompt for this is shown in Figure \ref{FinGrAct_error_seg}.

\begin{figure}
\fcolorbox{blue}{lime}{%
        \begin{minipage}{\columnwidth}
\scriptsize
You will receive a claim, and some trustworthy reliable information called evidence.  Your task is to divide the claim into multiple smaller subclaims/ atomic claims
  , then assess the factuality of each subclaim sentence based on the information provided in the evidence:\\

- no error: the subclaim aligns explicitly with the content of the evidence and is factually consistent with it.\\
- factuality error: the subclaim contains any factuality error.
\\\\
Instruction:\\
First, compare each subclaim sentence with the evidence.
Second, provide a single sentence explaining the factuality error in the subclaim and how to correct it based on the evidence.
\\\\
Provide your answer in JSON format. Your answer should strictly be a list of dictionaries whose keys are "sentence", "reason" and "correction". An example of your output should be:

[\{"sentence": "first  subclaim", "reason": "your reason", "correction": "your correction"\},\\
\{"sentence": "second  subclaim", "reason": "your reason", "correction": "your correction"\}]\\

Claim:\\
\{claim\}
\\
Evidence:\\
\{evidence\}
 \end{minipage}%
}
    \caption{FinGrAct prompt for error segmentation. The output should include false atomic claims, their factual errors, and GPT-4's proposed corrections. The explanation will then be evaluated to check if it explicitly covers all detected errors and corrections.}
    \label{FinGrAct_error_seg}

\end{figure}

After detecting errors and generating corrections, the next step is to verify whether these elements are explicitly inferred from the explanation. This involves answering “yes” or “no” for each detected error and proposed correction to assess alignment with the explanation.

The goal is to determine whether:

\begin{enumerate}
    \item The number of false atomic claims matches the number of errors mentioned in the explanation.
    \item The number of detected corrections corresponds to the number of corrections provided in the explanations.
\end{enumerate}
This process helps establish whether error detection and correction were partial or complete, which ultimately influences the final actionability score.

Continuing on the previous example of the claim that Earth is flat and red. Here is an example of the output:
[
\{'error': 'The evidence explicitly states that Earth is a marble shaped planet, not flat', 'response': 'Yes', 'correction': 'Yes', 'supporting\_links': 'Yes'\}, \{'error': 'As per the evidence, Earth is blue.', 'response': 'Yes', 'correction': 'Yes', 'supporting\_links': 'Yes'\}
]

The "error" key in the JSON output represents a specific factual error identified during the error segmentation and correction stage. The "response" key is a boolean indicating whether the explanation explicitly mentions the identified error. The "correction" key is another boolean that shows whether the explanation includes the corresponding correction from the error segmentation and correction process. Finally, the "supporting\_link" key is a boolean that signifies whether there is at least one link in the explanation with content that supports the correction, as assessed based on the LLM’s prior knowledge.

Additionally, each correction in the explanation—mapped to a correction identified during the error segmentation and correction phase in FinGrAct—is verified to ensure it has at least one supporting link with relevant web content. An explanation is deemed fully actionable only if at least one link supports all corrections.

Sanity checks were implemented to prevent impossible scenarios. For example, the "related\_links" key cannot be "yes" if "existing\_link" is not, and the "supporting\_link" boolean cannot be "yes" unless both "existing\_links" and "related\_links" are also "yes".

There are two methods for checking link content:

\begin{enumerate}
    \item LLM’s Prior Knowledge: The evaluator relies on the LLM’s pre-existing knowledge from training to assess whether the link content aligns with the corrections. The corresponding prompt is shown in Figure \ref{FinGrAct_act_eval}.
    \item URL Content Retrieval (UCR): The URL content retriever fetches the text content from the web page, which is then checked for alignment with the corrections. The corresponding prompt is shown in Figure \ref{FinGrAct_act_eval_with_lnk}.
\end{enumerate}

\begin{figure}
\fcolorbox{blue}{lime}{%
        \begin{minipage}{\columnwidth}
\scriptsize
You will receive a list of errors, their corrections, and a transcript called the 'explanation'. Your task is to assess if each of the errors can be inferred from the explanation, and if the corrections can be inferred from the explanation as well.

Instruction:\\\\
First, compare each error with the explanation.\\
Second, check if the error is inferred from the explanation and then respond "Yes" or "No" for each detected error explicitly mentioned in the explanation.\\
Third, compare each correction with the explanation.\\
Fourth, check if the correction is inferred from the explanation and then respond with "Yes" or "No" for each detected correction.\\
Fifth, based on your knowledge, check if there are credible and relevant web links in the explanation supporting the correction, and then respond with "Yes" or "No".\\\\

  Provide your answer in JSON format. The answer should be a list of dictionaries whose keys are "error", "response", "correction", and "supporting\_links". An example of your output:

  [
  \{"error": "first error", "response": "Yes", "correction": "Yes", "supporting\_links": "Yes"\},\\
  \{"error": "second error", "response": "No", "correction": "Yes", "supporting\_links": "No"\}, \\
  \{"error": "third error", "response": "Yes", "correction": "No", "supporting\_links": "No"\}
  ]\\\\
List of errors:\\
\{error\_list\}
\\\\
Corrections:\\
\{corrections\_list\}
\\\\
Explanation:\\
\{explanation\}
  
 \end{minipage}%
}
    \caption{FinGrAct Prompt for evaluating the actionability aspects. This prompt represents the actionability evaluation stage.}
    \label{FinGrAct_act_eval}

\end{figure}

\begin{figure}
\fcolorbox{blue}{lime}{%
        \begin{minipage}{\columnwidth}
\scriptsize
You will receive a list of errors, their corrections, and a transcript called the 'explanation'. Your task is to assess if each of the errors is can be inferred from the explanation, and if the corrections can be inferred from the explanation as well.\\\\

Instruction:\\\\
First, compare each error with the explanation.\\
Second, check if the error is inferred from the explanation and then response "Yes" or "No" for each detected error explicitly mentioned in the explanation.\\
Third, compare each correction with the explanation.\\
Fourth, check if the correction is inferred from the explanation and then respond with "Yes" or "No".\\
Fifth, check if there are working web links in the explanation. The links content will mention if the link is working or not, and then respond with "Yes" or "No".\\
Sixth, check the provided content of these web links, if the content is related, and then respond with "Yes" or "No".\\
Seventh, check the provided content of these web links, if the content supports the correction, and then respond with "Yes" or "No".\\
\\\\
  Provide your answer in JSON format. The answer should be a list of dictionaries whose keys are "error", "response", "correction", and "supporting\_links".\\
  An example of your output:
  [\{"error": "first error", "response": "Yes", "correction": "Yes", "existing\_links": "Yes", "related\_links": "Yes", "supporting\_links": "Yes"\},\\
   \{"error": "second error", "response": "No", "correction": "Yes", "existing\_links": "Yes", "related\_links": "No", "supporting\_links": "No"\},\\
   \{"error": "third error", "response": "Yes", "correction": "No",  "existing\_links": "Yes", "related\_links": "Yes", "supporting\_links": "No"\}]
\\\\
List of errors:\\
\{errors\_list\}
\\\\
Corrections:\\
\{corrections\_list\}
\\\\
Explanation:\\
\{explanation\}
\\\\
Links content:\\
\{links\_content\}
 \end{minipage}%
}
    \caption{FinGrAct Prompt for evaluating the action-
ability aspects. This prompt represents the actionability
evaluation stage and the source evaluation stage.}
    \label{FinGrAct_act_eval_with_lnk}

\end{figure}


\end{document}